\documentclass[]{spie}

\makeatletter
\def\endthebibliography{%
  \def\@noitemerr{\@latex@warning{Empty `thebibliography' environment}}%
  \endlist
}
\makeatother

\usepackage{amsmath,amsfonts,amssymb}
\usepackage{graphicx}
\usepackage[colorlinks=true, allcolors=blue]{hyperref}
\usepackage{caption}
\usepackage{subcaption}
\usepackage{float}

\title{Evaluation of neural network algorithms for atmospheric turbulence 
mitigation}

\author[a]{Tushar Jain}
\author[a]{Madeline Lubien}
\author[a]{J\'er\^ome Gilles}
\affil[a]{San Diego State University - Department of Mathematics and 
Statistics, 5500 Campanile Dr, San Diego, CA 92182, USA}

\authorinfo{Further author information: (Send correspondence to J\'er\^ome 
Gilles)\\J\'er\^ome Gilles: E-mail: jgilles@sdsu.edu, Telephone: 1 619 594 
7240}

\begin{document} 
\maketitle

\begin{abstract}
    A variety of neural networks architectures are being studied to tackle blur 
in images and videos caused by a non-steady camera and objects being captured. 
In this paper, we present an overview of these existing networks and perform 
experiments to remove the blur caused by atmospheric turbulence. Our experiments 
aim to examine the reusability of existing networks and identify desirable 
aspects of the architecture in a system that is geared specifically towards 
atmospheric turbulence mitigation. We compare five different architectures, 
including a network trained in an end-to-end fashion, thereby removing the need 
for a stabilization step.
\end{abstract}


\section{INTRODUCTION}
Atmospheric turbulence mitigation has attracted a lot of attention these last 
few decades with the achievement of long-range sensors (see for instance 
Ref.~\citenum{Mugnier2008,Schutte2012,Yao2016,Huebner2016} for some general 
reviews). The optical path going 
through the atmospheric is altered by the inhomogeneity of the 
refractive index created by the presence of turbulence. This results in the 
appearance of blur as well as geometric distortions of the image (the blur is 
commonly considered stationary compared to the geometric distortions, i.e. 
among several frames acquired within a few seconds, the blur does not change 
with time). A widely accepted general image formation model is given by
$$f_i=\Phi_i(H\ast u)+n_i,$$
where $f_i$ is an observation at time $i$, $\Phi_i$ represents the geometric 
distortions induced at time $i$, $H$ corresponds to the transfer function of 
the atmosphere and imaging system combined inducing blur in images, $n_i$ is 
some noise, and $u$ is the underlying clean image we expect to recover. 
Inverting the operator $H$ corresponds to a deconvolution problem while 
inverting the sequence of operators $\Phi_i$ is usually seen as a 
stabilization/unwarping problem. Many different approaches have 
been proposed to address each step.\\ 
Stabilization is often modeled using 
elastic registration techniques. First, the deformation fields, $\Phi_i$, 
between each frame $f_i$ and a reference image (usually obtained using some 
temporal filtering like mean or median) are estimated. This information is then
used to compensate the local geometric deformations. Optical 
flow based 
variational 
techniques~\citenum{Fraser2003,Gepshtein2004a,Zamek2006,Shimizu2008,Mao2012c, 
Eekeren2012,Yang2013,Gilles2013b,Gal2014a,Song2015,Huebner2016,Xie2016,
Nieuwenhuizen2017,Nieuwenhuizen2019}, block matching type 
approaches~\citenum{Huebner2008,Huebner2009,Huebner2010,Halder2013,Halder2013a, 
Anantrasirichai2013,Huebner2016}, or control 
points/grid~\citenum{Frakes2001,Zhu2010,Zwart2012,He2016} are the most common 
used techniques to estimate the $\Phi_i$. They are then followed by some 
interpolation technique (like 
bilinear~\citenum{Frakes2001,Fraser2003,Mao2012c}, 
sinc~\citenum{Gepshtein2004a}, 
B-spline~\citenum{Unser1989a,Shimizu2008,Zhu2010,Halder2013a,Yang2013,Zhu2013, 
He2016,Furhad2016}) to compute the final registration. Diffeomorphic mappings, 
as well as dynamical systems approaches have been considered 
in~\citenum{Gilles2008} and \citenum{Micheli2013}, respectively. Starting 
from the assumption that, across time, some local 
sharp version of the pristine image appears, image fusion techniques have also 
been investigated. The idea is to appropriately 
select patches of the image across space and time and to fusion them together to 
reconstruct the restored image. The two most used fusion methods are the 
lucky-imaging~\citenum{Aubailly2009,Schutte2012,Anantrasirichai2013,Yang2013, 
Lou2013,Mao2020} and centroid~\citenum{Mario2012,Halder2013,Micheli2013,Enric2014} 
techniques, respectively.\\
In turbulence mitigation, the deconvolution step is usually challenging since 
the kernel $H$ is unknown. This leads to either use blind or semi-blind (i.e a 
model for $H$ is available where it is only needed to estimate its parameters) 
deconvolution techniques. Modeling the kernel $H$ has been addressed 
in~\citenum{Sadot1995,Schulz1997,Bondeau1998,Xu1998a,Cochran1999,Li2005,
Iersel2010,
Droege2012,Gilles2012,Deledalle2020,Gao2022}, either combining optics laws with 
turbulence equations; or finding simpler generic models corresponding to 
measurements. Equipped with such kernel models, different semi-blind 
deconvolution methods were proposed: Maximum-Likelihood~\citenum{Schulz1997}, 
Thikonov type minimization~\citenum{Cochran1999,Lemaitre2005a,Lemaitre2006a}, 
Wiener 
filtering~\citenum{Denker2004a,Li2005,Lemaitre2005a,Lemaitre2006a,Huebner2008,
Li2009,Iersel2010,Gal2014a}, 
Principal Component Analysis~\citenum{Li2007,Huebner2008,Lu2021}. The most 
popular 
blind deconvolution algorithms used in the atmospheric mitigation literature are: 
Lucy-Richardson~\citenum{Huebner2008,Huebner2009,Iersel2010,Gal2014a}, 
least-square 
minimization~\citenum{Fraser2003,Hong2012}, iterative 
methods~\citenum{Fraser2003,Huebner2008,Zuo2009,Huebner2009,Huebner2010,
Huebner2012,Holmes2019}, multi-frame blind 
deconvolution~\citenum{Denker2004a,Gilles2016a,Hajmohammadi2019}, Maximum 
Likelihood~\citenum{Lemaitre2006a,Chunsheng2006,Hameling2010,Song2015,He2016}, and
variational 
models~\citenum{Gilles2008,Hirsch2010,Zhu2010,Wang2011,Gilles2012,Eekeren2012,
Mario2012,Zhu2013,Micheli2013,Gal2014a,Xie2016,Furhad2016,Deledalle2020,Duan2021
}.\\
These last years, neural networks have became very popular to solve computer 
vision tasks, and in particular to perform deconvolution. Surprisingly, their 
use for turbulence mitigation has not been seriously investigated at the time 
we write this article. The only two available references simply use existing 
deconvolution neural network architectures to perform the last deblurring step. 
In \citenum{Nieuwenhuizen2019}, the authors implement a denoising convolutional 
neural network (DnCNN), while the authors of~\citenum{Bai2019} investigate 
the use of either a fully convolutional network (FCN) or a conditional 
generative adversarial network (CGAN) to mitigate the blur (these approaches 
follow an initial stabilization step as the ones described above). To the best 
of our knowledge, no neural networks have been built to specifically address 
the turbulence mitigation problem. This is probably due to the fact that 
such neural networks need a very large amount of data (including the 
groundtruths) for their training, and that such large open dataset is currently 
not available and difficult/expensive to acquire. To remediate this 
question and open the door to the construction of such dedicated neural network 
architectures, we propose the creation of a publicly open dataset, called SOTIS, 
using a realistic turbulence simulator.
Given the vast literature on turbulence mitigation, it also becomes crucial to 
propose some evaluation methodology to compare different algorithms. Such 
evaluation aspects have been very quickly addressed 
in~\citenum{Huebner2008,Espinola2012,Eekeren2014,Gal2014a,Huebner2016,
Kozacik2017,Yang2018a} and we also propose to define such process.\\
The reminder of the paper is organized as follow. Section~\ref{sec:eval} is 
devoted to the creation of the simulated dataset and the definition of the 
evaluation methodology. Section~\ref{sec:deep} provides some background on the 
different neural network architectures we will use to perform the deblurring. 
Experimental results on both classic algorithms and deep learning based 
algorithms will be provided in Section~\ref{sec:exp}. Finally, this work will 
be concluded in Section~\ref{sec:conc}.

\section{Performance evaluation: dataset and metrics}\label{sec:eval}
In this section, we briefly describe the process we use to create 
the Simulated Open Turbulence Image Set (SOTIS). We, then, define the 
evaluation method we propose to the community to assess turbulence mitigation 
algorithms performances.
\subsection{Simulated dataset}
\begin{figure}[!t]
 \begin{tabular}{ccc}
  \includegraphics[width=0.3\textwidth]{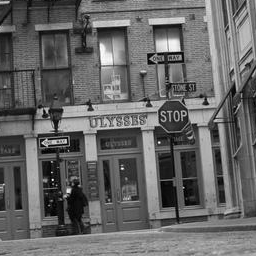} &
  \includegraphics[width=0.3\textwidth]{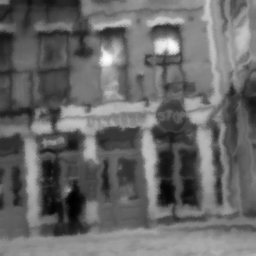} &
  \includegraphics[width=0.3\textwidth]{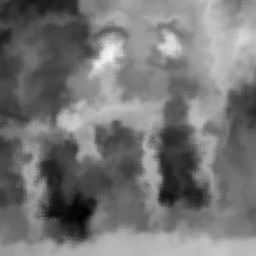} \\
  \includegraphics[width=0.3\textwidth]{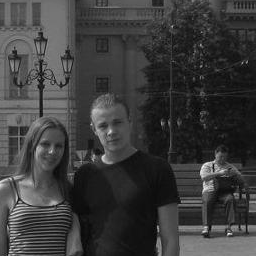} &
  \includegraphics[width=0.3\textwidth]{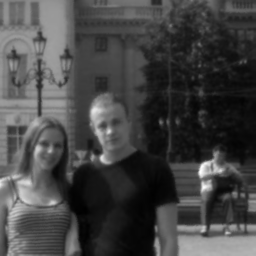} &
  \includegraphics[width=0.3\textwidth]{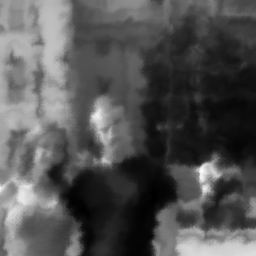} \\
  \includegraphics[width=0.3\textwidth]{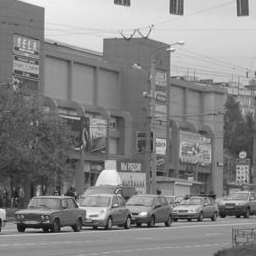} &
  \includegraphics[width=0.3\textwidth]{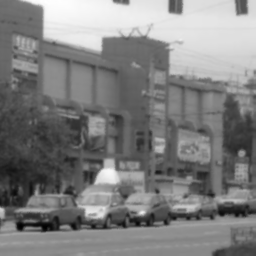} &
  \includegraphics[width=0.3\textwidth]{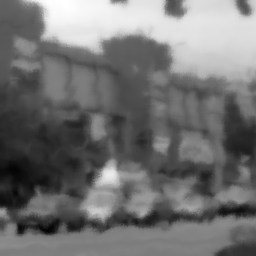} \\
 \end{tabular}
\caption{Examples of available sequences in the SOTIS dataset. The used ground-truth images are given in the left column. 
    The corresponding weak and 
    strong turbulence scenarios are illustrated in the center and right columns, respectively.}
\label{fig:sotis}
\end{figure}

Acquiring images through the atmospheric turbulence is a time consuming and 
expensive process, not mentioning that we can't control the turbulence strength 
during these acquisitions. However, with the availability of deep learning 
algorithms, the need for large datasets is becoming critical. To train such 
algorithms, as well as to perform quantitative evaluations, ground-truth images
are also needed, which are never available while doing 
acquisition on the field as they would correspond to have no turbulence. It is 
for all these reasons that atmospheric turbulence simulation tools have been 
studied in the 
literature~\citenum{Pigois2004,Thomas2004,Repasi2011,Huebner2012,Rampy2012,Chimitt2020}. 
The most popular approach to simulate turbulence is based on the generation of random phase
screens. To create the SOTIS dataset, we used the method described in~\citenum{Chimitt2020},
its MATLAB code is available on the authors 
website~\footnote{\url{https://engineering.purdue.edu/ChanGroup/index.html}} (we only modified 
the code to incorporate some parallelization to speed up the process). The simulation algorithm 
generates anisotropic random Zernike phase screens which are then used to distort the images. 
Long and short exposures, as well as spatial correlations are also taken into account, providing
a very realistic simulator.\\
To guarantee enough variability in our dataset (i.e. images containing buildings, pedestrians, 
vehicles, vegetation, signs,\ldots), we cropped 215 images of size 256$\times$256 to serve as 
ground-truth images from the Eurasian-Cities dataset~\citenum{Tretiak2012}. The simulator main 
parameters are: the focal length $d=0.3$, the aperture diameter $D=0.054$ (we kept these values
to the default ones given by the authors of~\citenum{Chimitt2020}), the distance sensor/scene $L$,
the refractive index $C_n^2$. To create a wide range of turbulence and observation scenarios, we 
sample $L$ and $C_n^2$ as follows: $L=1,2,3,4$ km, $C_n^2=a^{-b}$ where $a=1,3,5,7,9$ and $b=14,15,16,17$.
We fix the number of frames in the generated sequences to $N=50$. These choices lead to the creation
of 80 sequences for each ground-truth, resulting in a total of 17400 sequences in SOTIS. Figure~\ref{fig:sotis},
illustrates some frame examples from the available sequences in both weak and strong turbulence cases. 
The SOTIS dataset is made publicly available, all information can be found on the third author's webpage\footnote{https://jegilles.sdsu.edu/datasets.html}.

\subsection{Evaluation protocol}
Given the large literature on atmospheric turbulence mitigation, it becomes imperative to define some evaluation protocol 
to fairly and quantitatively compare the different algorithms. Algorithm evaluation has been considered in~\citenum{Huebner2008,Eekeren2014,Gal2014a,Huebner2016,Kozacik2017,Yang2018a}. However, these articles do not really define 
a specific protocol, and mostly struggle of achieving their goal because of the lack of ground-truth images in most cases.\\
Since SOTIS provides the ground-truths images (which will be denoted $f_{gt}$ hereafter), we propose to use the widely accepted peak signal to 
noise ratio (PSNR) and structural similarity index measure (SSIM) metrics respectively defined by (we denote $f_{rest}$ the 
restored image):
$$PSNR(f_{gt},f_{rest})=10\log_{10}\left(\frac{\max_{im}}{\|f_{gt}-f_{rest}\|_2^2}\right),$$
where $\max_{im}$ is the maximum value an image can reach (255 for 8-bits encoded images); and
$$SSIM(f_{gt},f_{rest})=\frac{(2\mu_{gt}\mu_{rest}+c_1)(2\sigma_{gt,rest}+c_2)}{(\mu_{gt}^2+\mu_{rest}^2+c_1)(\sigma_{gt}^2+\sigma_{rest}^2+c_2)},$$
where $\mu_{gt}$ and $\mu_{rest}$ are respectively the average of $f_{gt}$ and $f_{rest}$; $\sigma_{gt},\sigma_{rest},\sigma_{gt,rest}$ their variances and covariance, $c_1,c_2$ two constants defined from the images dynamic range. Notice that the SSIM metric provides values in the range $[0,1]$ (1 being the best performance).\\
With the SOTIS dataset, we provide several MATLAB scripts where the user can easily plug any stabilization, deblurring or combined 
algorithms and create the appropriate directory structure to store his results. We also provide a script that parses all the 
results and build a CSV file that contains the corresponding PSNR and SSIM values. Any statistical software like R can then extract 
all the useful evaluation statistics (we also provide the R script we wrote to generate the results in this article).

\section{Deep learning and turbulence mitigation}\label{sec:deep}
Neural networks have gained a lot attention towards recovering images degraded by blur caused by camera shake. These networks use a variety of techniques (we refer the reader to the review paper~\citenum{KOH2021103134}), and particularly show significant success in dynamic scene deblurring; which incorporates blur caused by moving objects in addition to the blur due to camera shake. Such architecture inverts the effect of a blur kernel that is non-uniform, i.e. it varies from pixel to pixel. Because atmospheric turbulence induced blur can be seen as such anisoplanatic blur, we evaluate the three best ranked architectures from~\citenum{KOH2021103134} on the SOTIS dataset. Furthermore, we chose two other models that were not mentioned in~\citenum{KOH2021103134}, in order to analyze different neural networks architectures. Most neural networks for deblurring are designed to process single images, therefore we use the stabilization results as our  input for these neural networks. We also try end-to-end training on a neural network that aims to achieve video deblurring and takes an entire sequence of images as input, thereby forgoing the need of the stabilization step. We consider the following networks for evaluation,

\begin{itemize}
  \item Scale-recurrent Network for Deep Image Deblurring (SRN)~\citenum{DBLP:journals/corr/abs-1802-01770}
  \item Deep Multi-scale Convolutional Neural Network for Dynamic Scene Deblurring (DSD)~\citenum{DBLP:journals/corr/NahKL16}
  \item DeblurGAN-v2 (DGv2)~\citenum{DBLP:journals/corr/abs-1908-03826}
  \item Deblurring using Analysis-Synthesis Networks Pair (ASD)~\citenum{DBLP:journals/corr/abs-2004-02956}
  \item Cascaded Deep Video Deblurring Using Temporal Sharpness Prior (CDVD-TSP)~\citenum{DBLP:journals/corr/abs-2004-02501}
\end{itemize}

It is common for such neural networks to use a U-Net based architecture~\citenum{DBLP:journals/corr/RonnebergerFB15} along with residual layers. Increasing the receptive field plays a key role and has been emphasized by DSD, SRN and CDVD-TSP. However, each neural network has a different method of increasing the receptive field. Apart from the multi-scale information that is extracted by the U-Net, an extra multi-scale pyramid decomposition is used in SRN by applying a U-Net at each resolution of the image. Then, the corresponding loss function is able to take into account the reconstruction at each specific resolution. The DSD architecture builds an IIR (Infinite Impulse Response) filter, while CDVD-TSP learns a temporal sharpness prior that is able to take into account long range dependency among the input sequence. Networks used for object detection and image segmentation, such as VGG16~\citenum{simonyan2015deep}, have shown to be useful for deblurring since they can identify regions of uniform blur. Such strategy is used in DGv2 and DSD to better estimate the blur kernel. The DSD and ASD networks have a separate kernel estimation module, whereas other networks do not separate the process of kernel estimation and image deblurring. The networks ASD and DGv2 introduced the use of novel concepts such as cross correlation layers, and adapting relativistic warping to the LS GAN~\citenum{DBLP:journals/corr/abs-1908-03826} loss.

\section{Experimental evaluations}\label{sec:exp}
In this section, we present some evaluation results for both classic (i.e non-deep learning) and deep-learning based mitigation algorithms. Every algorithms are based on an initial stabilization step followed by some deblurring.
\subsection{Non-Deep Learning case}
\begin{figure}[!t]
\centering
\begin{subfigure}{0.32\textwidth}
 \centering
 \includegraphics[width=\textwidth]{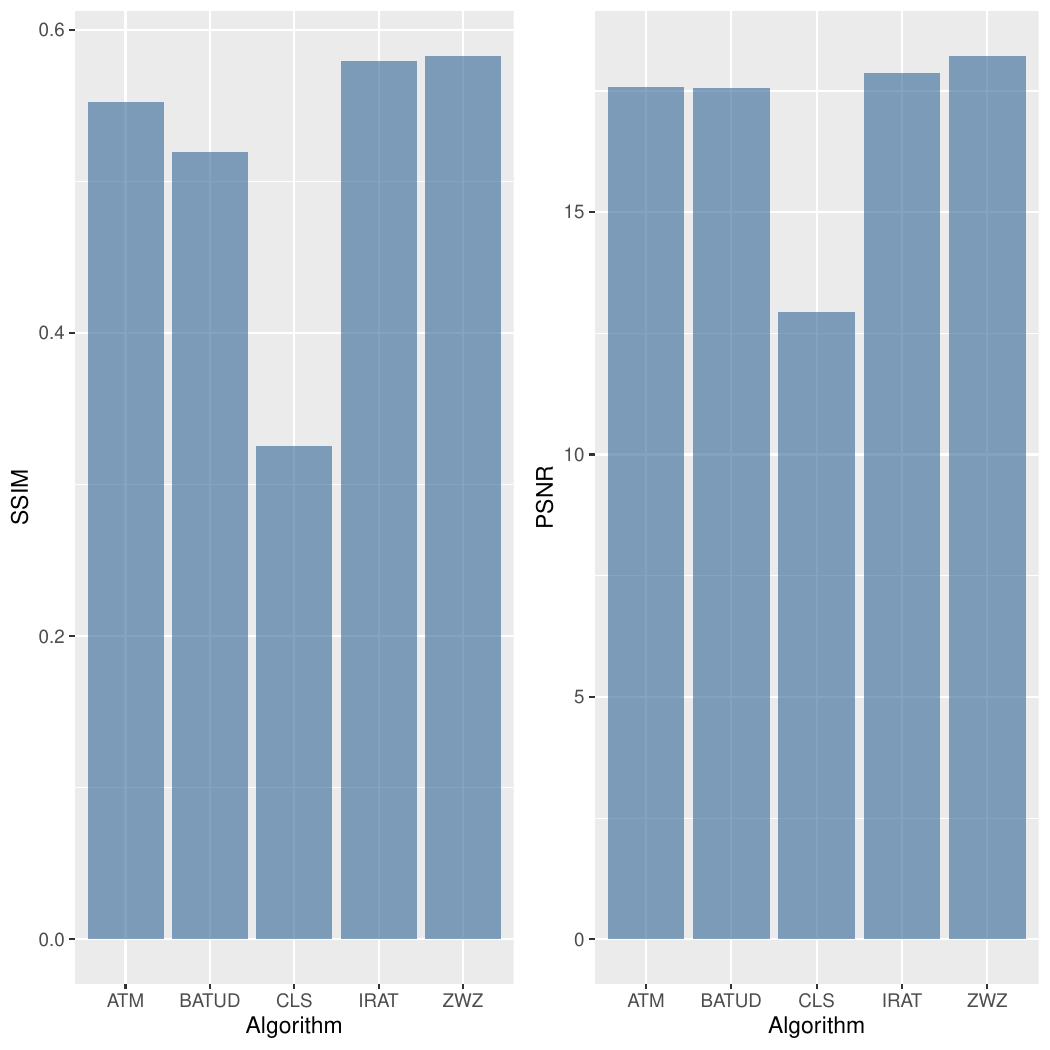}
 \caption{Overall performances (averaged across all stabilization algorithms).}
 \label{fig:ndl1}
\end{subfigure}
\hfill
\begin{subfigure}{0.32\textwidth}
 \centering
 \includegraphics[width=\textwidth]{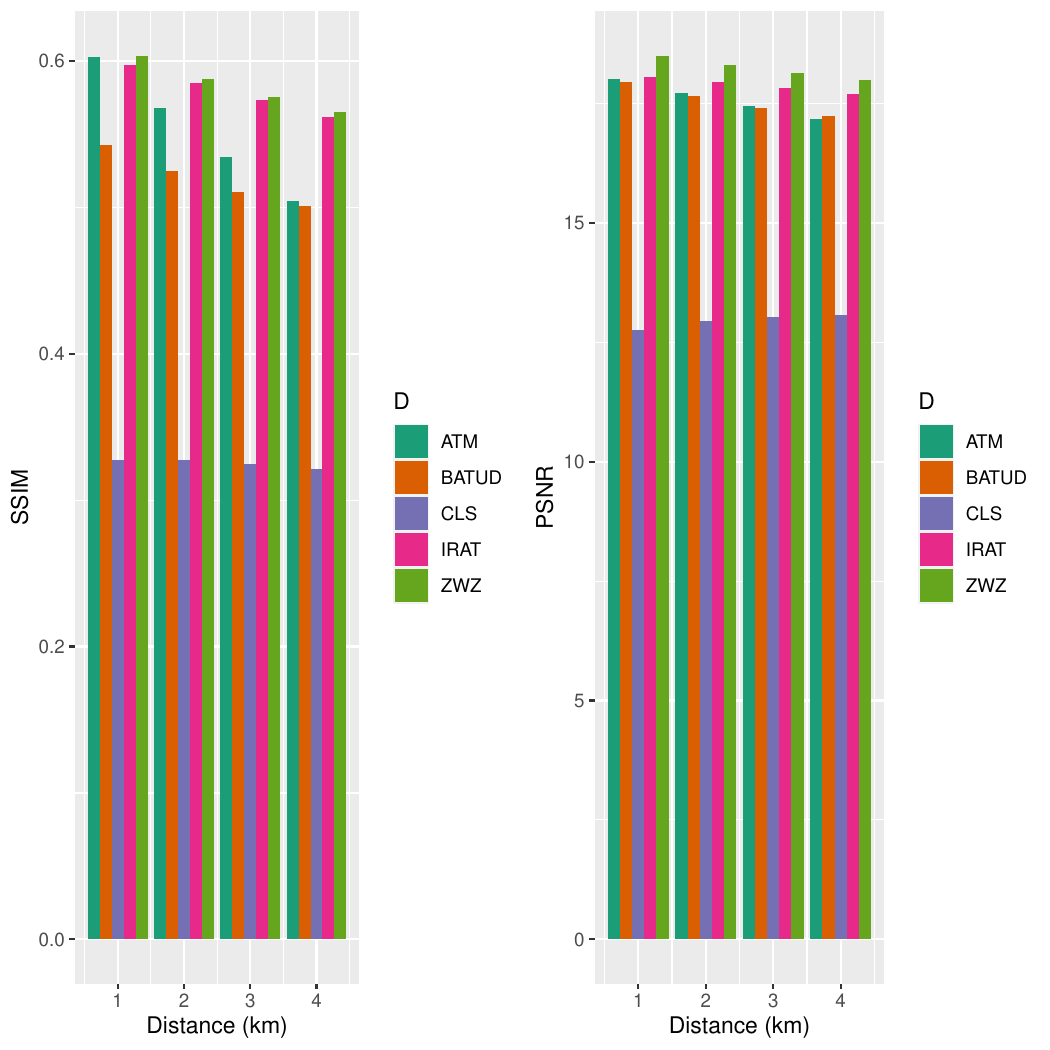}
 \caption{Performance with respect to the sensor-target distance $L$.}
 \label{fig:ndl2}
\end{subfigure}
\hfill
\begin{subfigure}{0.32\textwidth}
 \centering
 \includegraphics[width=\textwidth]{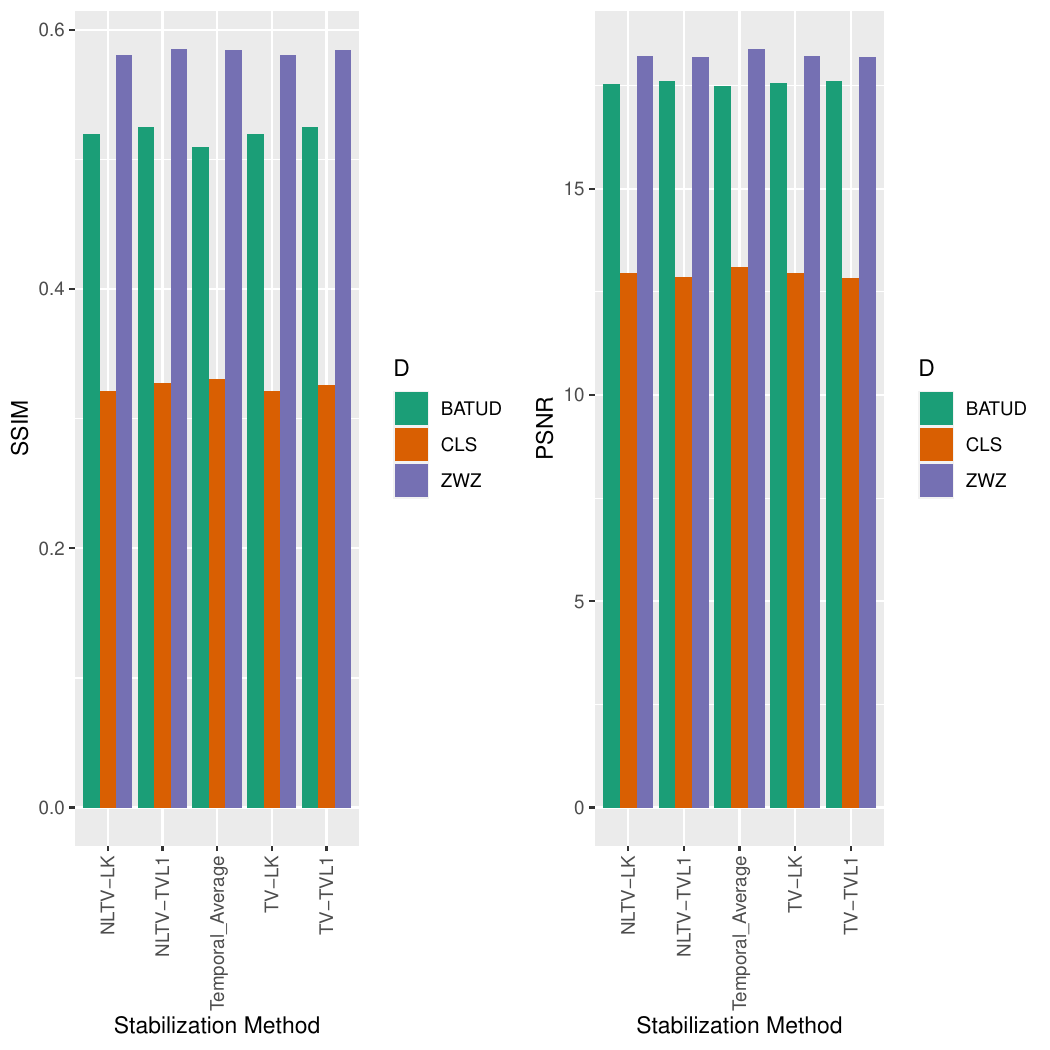}
 \caption{Performance with respect stabilization.}
 \label{fig:ndl3}
\end{subfigure}
\vspace{0.3in}

\begin{subfigure}{0.48\textwidth}
 \centering
 
 \includegraphics[width=\textwidth]{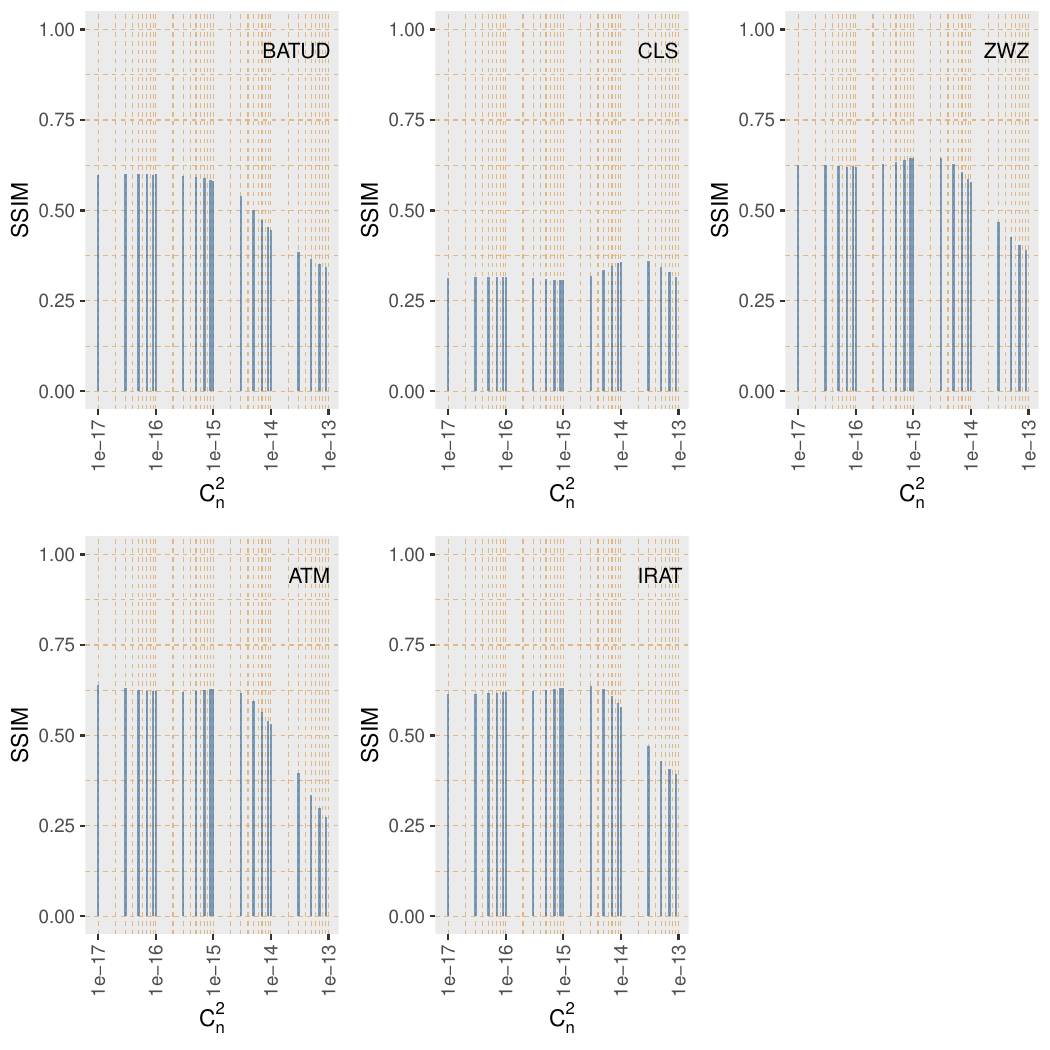}
 \caption{Performance (SSIM) with respect to turbulence strength ($C_n^2$).}
 \label{fig:ndl4}
\end{subfigure}
\hfill
\begin{subfigure}{0.48\textwidth}
 \centering
 \includegraphics[width=\textwidth]{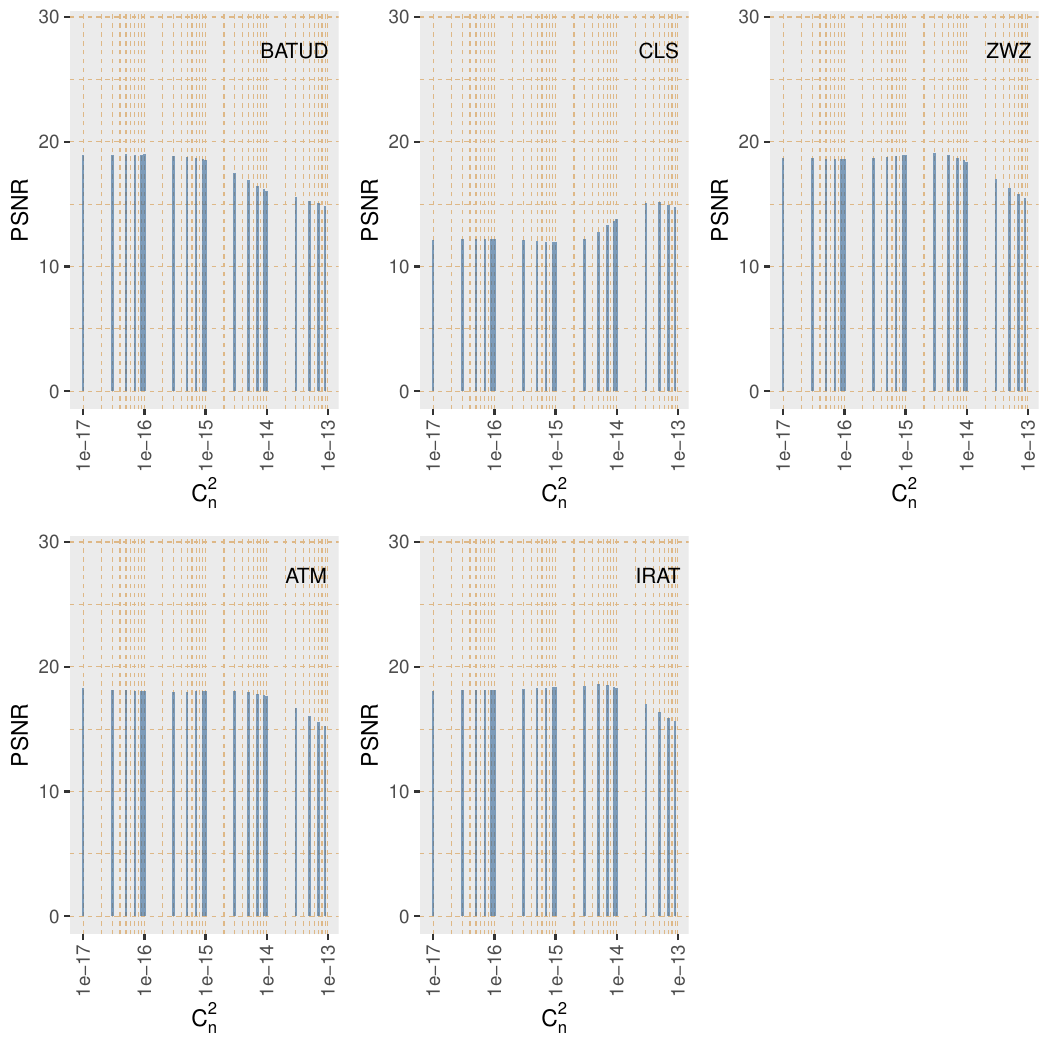}
 \caption{Performance (PSNR) with respect to turbulence strength ($C_n^2$).}
 \label{fig:ndl5}
\end{subfigure}
\vspace{0.3in}
\caption{Performances results for Non-Deep Learning algorithms.}
\label{fig:ndl}
\end{figure}

We test two main approaches for the stabilization: 1) a temporal average, 2) the Mao-Gilles~\citenum{Mao2012c} algorithm based on optical flow. For the later, we experiment the use of two different regularizations: $TV$ (total variation) or $NLTV$ (non-local total variation); and two variations of the optical flow: Lucas-Kanade or $TVL^1$ (total variation + $L^1$). These different stabilization methods will be denoted, Temporal\_Average, $TV-LK, TV-TVL1, NLTV-LK$ and $NLTV-TVL1$, respectively.\\
The used deblurring techniques are BATUD\footnote{\url{https://www.charles-deledalle.fr/batud}}~\citenum{Deledalle2020}, 
CLS\footnote{\url{https://blog.nus.edu.sg/matjh/files/2019/01/BlindDeblurSingleTIP-2jrncsd.zip}} (framelet based deconvolution)~\citenum{Cai2012}, and ZWZ\footnote{\url{https://drive.google.com/file/d/0BzoBvkfRHe5bUF9jQ1ZsWXRYSkk/edit?usp=sharing}} ($\ell^2-\ell_p$ sparse prior based variational model)~\citenum{Zhang2013}.\\
We also test the wavelet fusion based algorithm proposed by \citenum{Anantrasirichai2013}, denoted ATM\footnote{\url{https://github.com/pui-nantheera/atmospheric-turbulence-removal/}}, and the lucky imaging approach developed in~\citenum{Mao2020}, denoted IRAT\footnote{\url{https://github.itap.purdue.edu/StanleyChanGroup/TurbRecon_TCI}}. These algorithms are combined approaches, i.e they have their own stabilization and deconvolution approaches.\\
The evaluation results are given in Figure~\ref{fig:ndl}. Figure~\ref{fig:ndl1} provides the overall results, i.e averaged over all stabilization algorithms. If clearly the CLS algorithm does not perform well compared with the others, we can see that all other methods perform quite equivalently. It is interesting to notice that the SSIM metric always remains below 0.6, i.e. there is room to improve the existing algorithms. Figure~\ref{fig:ndl2} shows the influence of the distance between the target and the sensor. As expected, the further the target (hence the worse the degradation), the lower the performances. Figure~\ref{fig:ndl3} illustrates the impact of the chosen stabilization algorithm. Surprisingly, the simple temporal averaging performs as good as more advanced stabilization algorithms. This tends to confirm the idea, that a simple mean transforms the degradation in an isotropic blur that can be managed by efficient standard deconvolution algorithms. Finally, Figures~\ref{fig:ndl4} and \ref{fig:ndl5} give the performances with respect to the turbulence strength ($C_n^2$). Here again, as expected, the stronger the turbulence, the worse the degradation, and the lower the performances.

\subsection{Deep learning case}
\begin{figure}[!t]
  \centering
  \begin{subfigure}{0.32\textwidth}
   \centering
   \includegraphics[width=\textwidth]{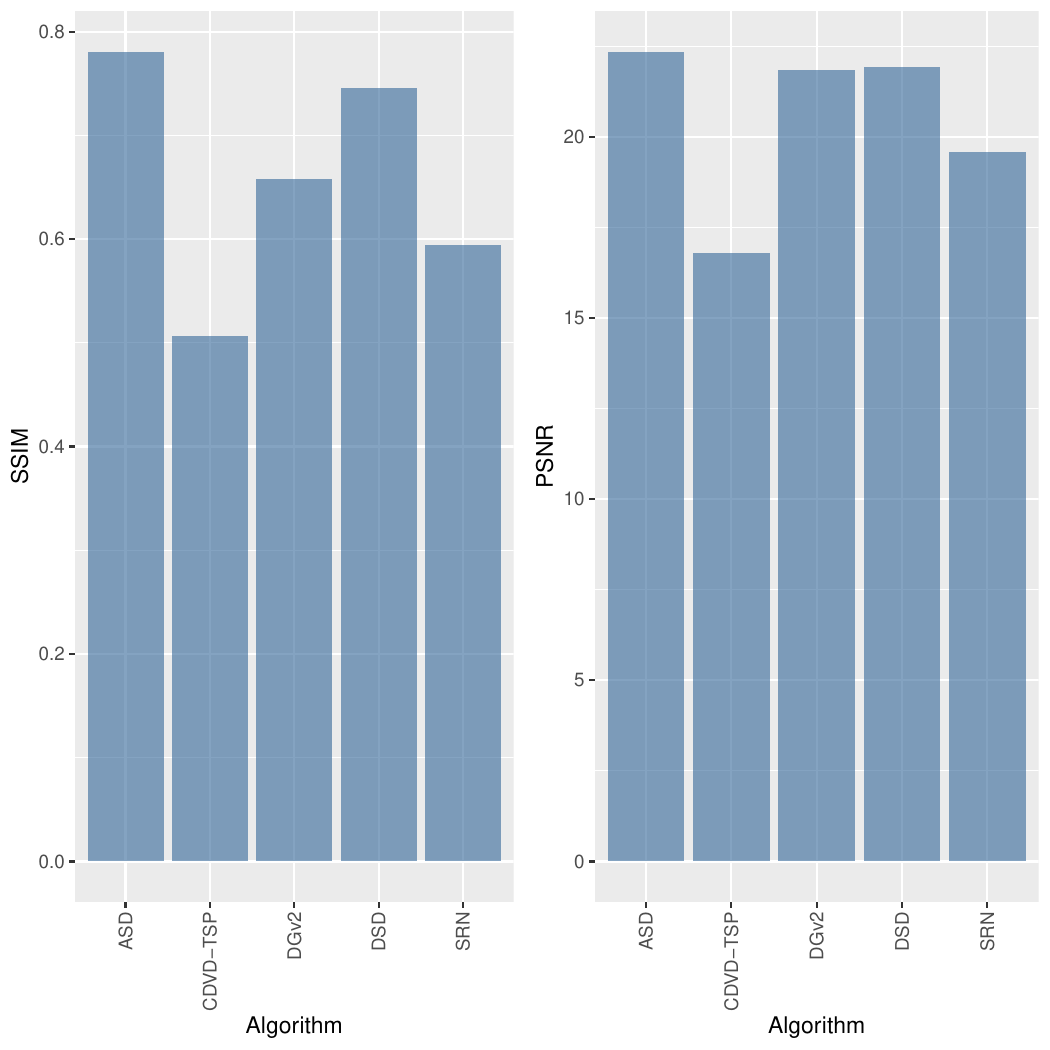}
   \caption{Overall performances (averaged across all stabilization algorithms).}
   \label{fig:dl1}
  \end{subfigure}
  \hfill
  \begin{subfigure}{0.32\textwidth}
   \centering
   \includegraphics[width=\textwidth]{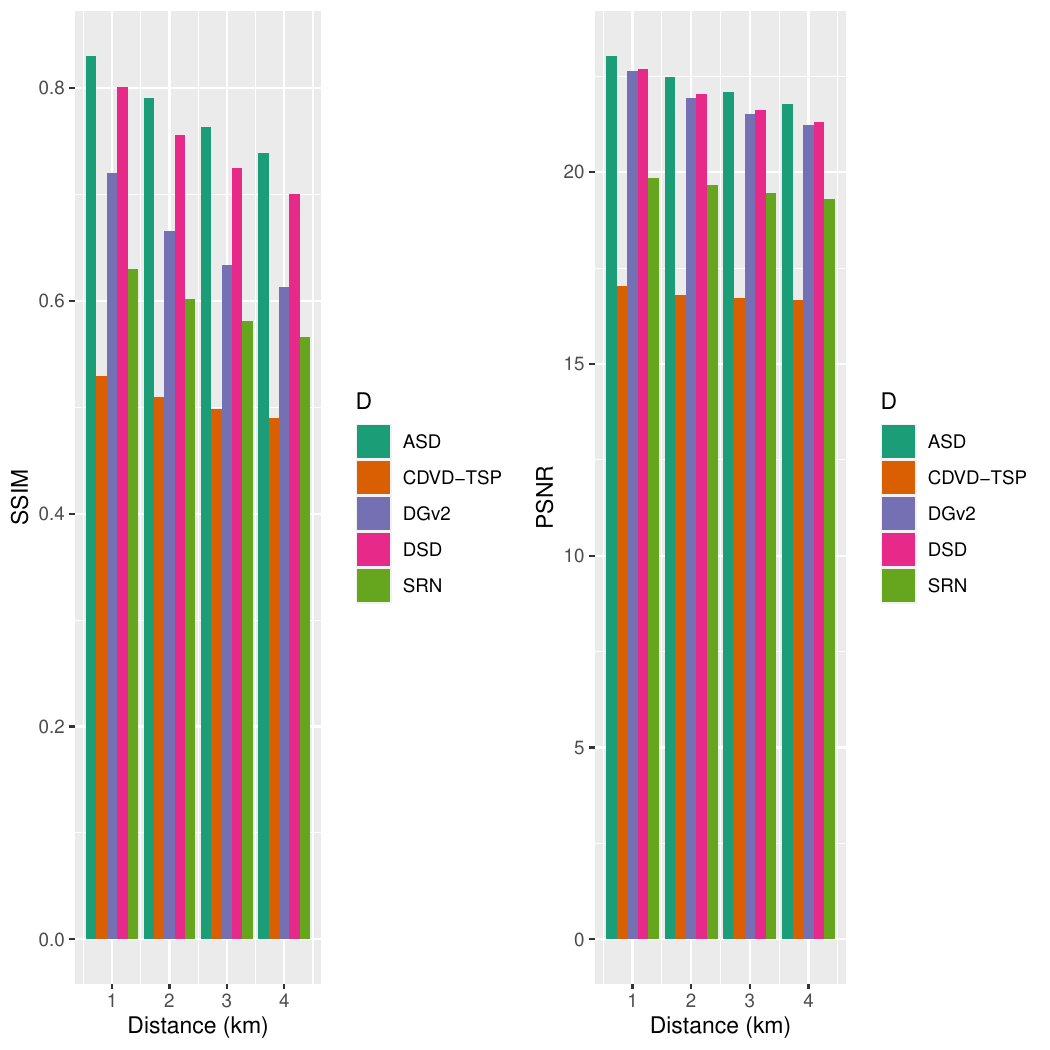}
   \caption{Performance with respect to the sensor-target distance $L$.}
   \label{fig:dl2}
  \end{subfigure}
   \hfill
   \begin{subfigure}{0.32\textwidth}
    \centering
    \includegraphics[width=\textwidth]{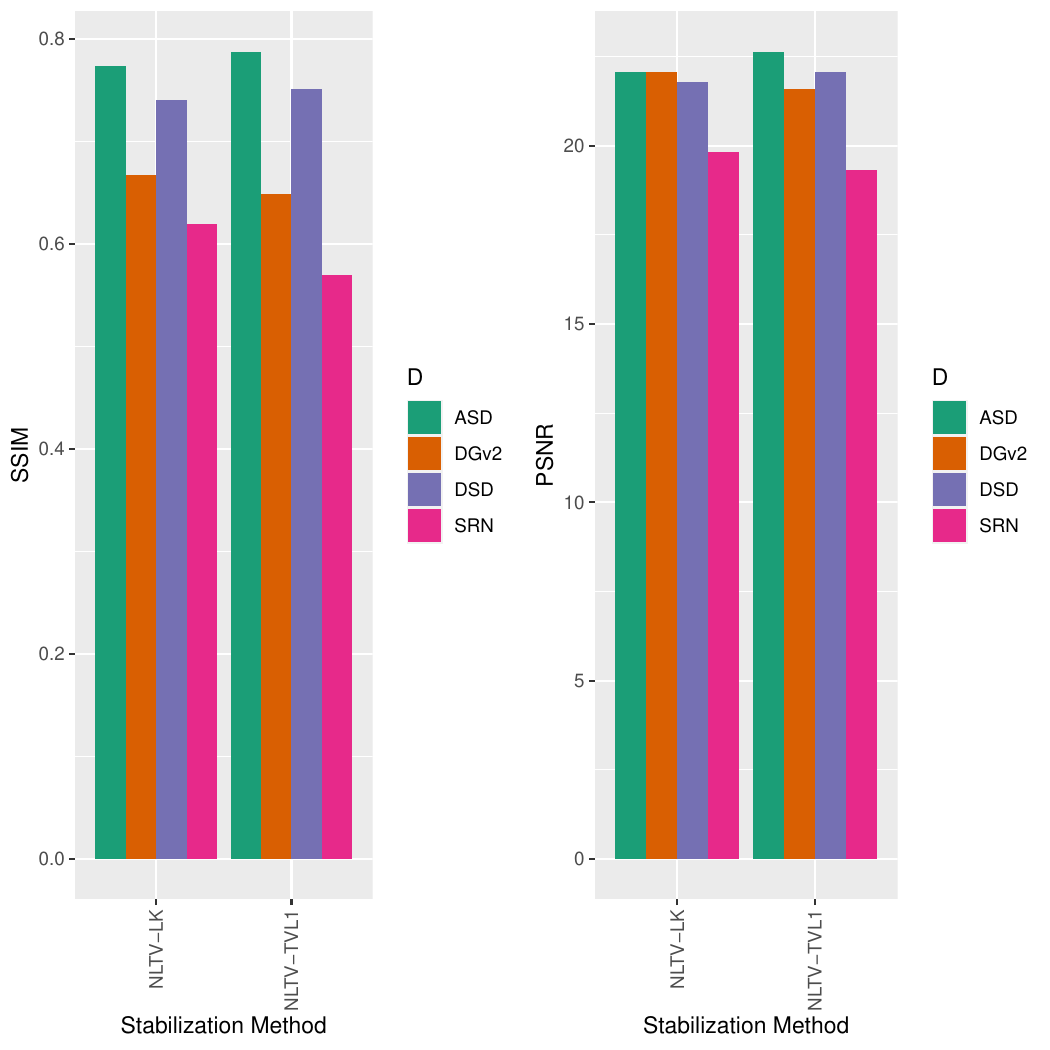}
    \caption{Performance with respect to stabilization.}
    \label{fig:dl3}
   \end{subfigure}
  \vspace{0.3in}
  
  \begin{subfigure}{0.48\textwidth}
   \centering
   \includegraphics[width=\textwidth]{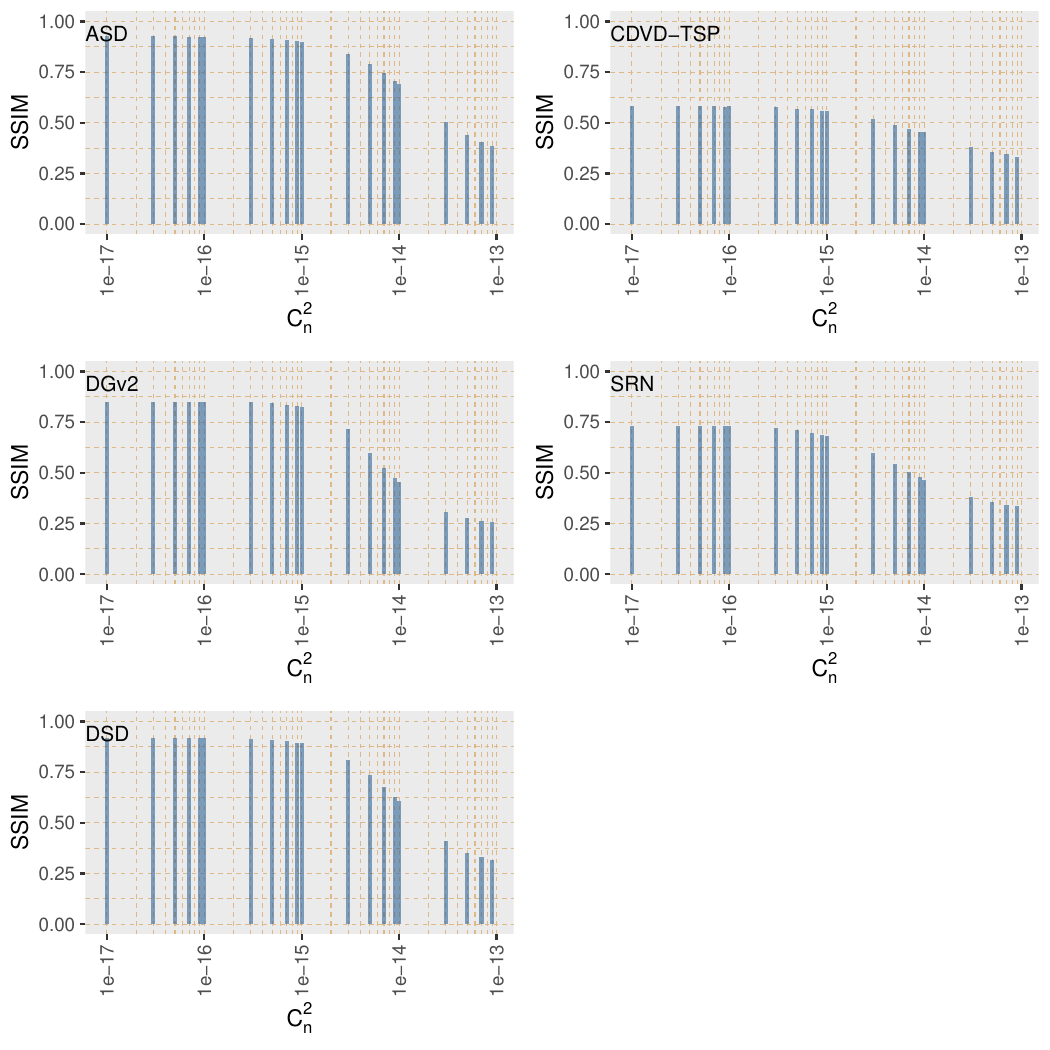}
   \caption{Performance (SSIM) with respect to turbulence strength ($C_n^2$).}
   \label{fig:dl4}
  \end{subfigure}
  \hfill
  \begin{subfigure}{0.48\textwidth}
   \centering
   \includegraphics[width=\textwidth]{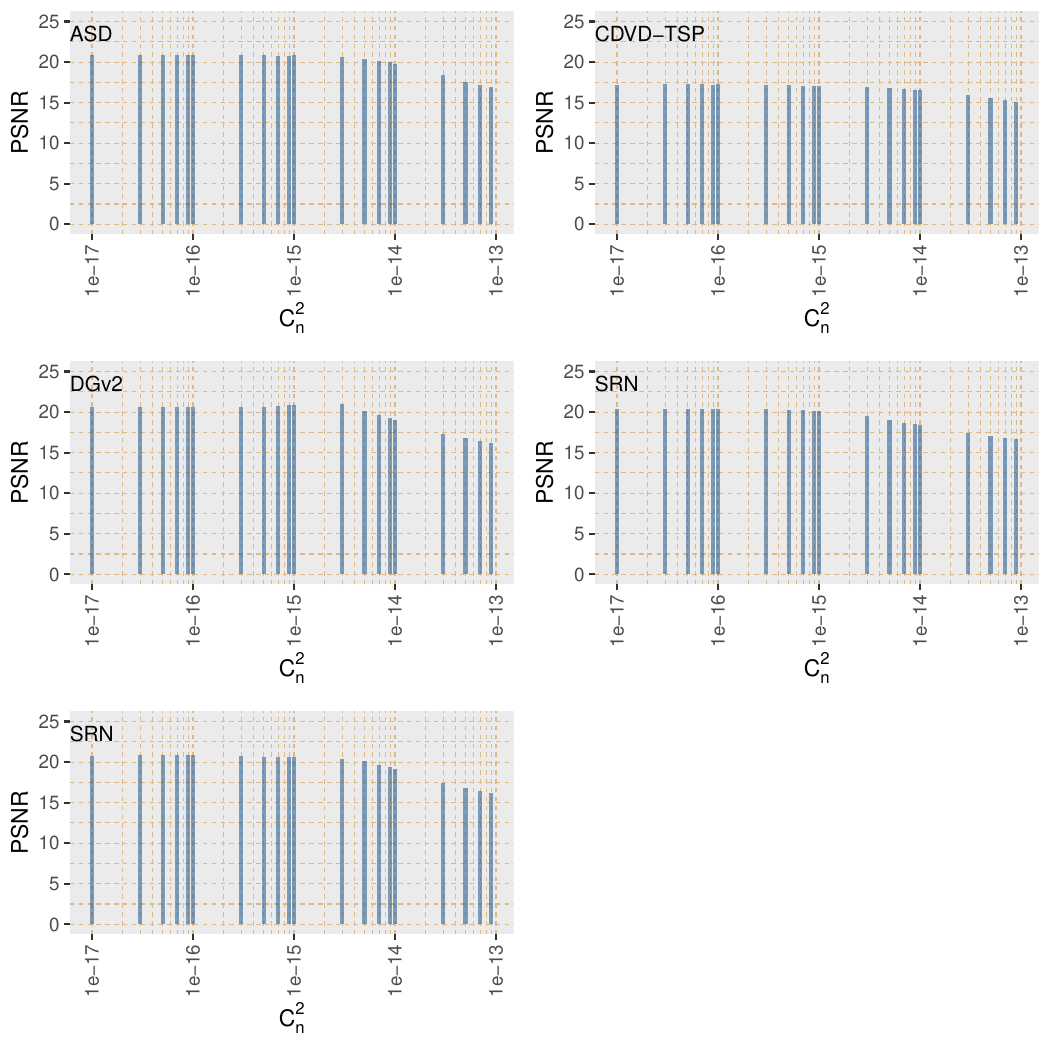}
   \caption{Performance (PSNR) with respect to turbulence strength ($C_n^2$).}
   \label{fig:dl5}
  \end{subfigure}
  \vspace{0.3in}
  \caption{Performances results for Re-trained Deep Learning algorithms.}
  \label{fig:dl}
  \end{figure}

\begin{figure}[!t]
  \centering
  \begin{subfigure}{0.32\textwidth}
   \centering
   \includegraphics[width=\textwidth]{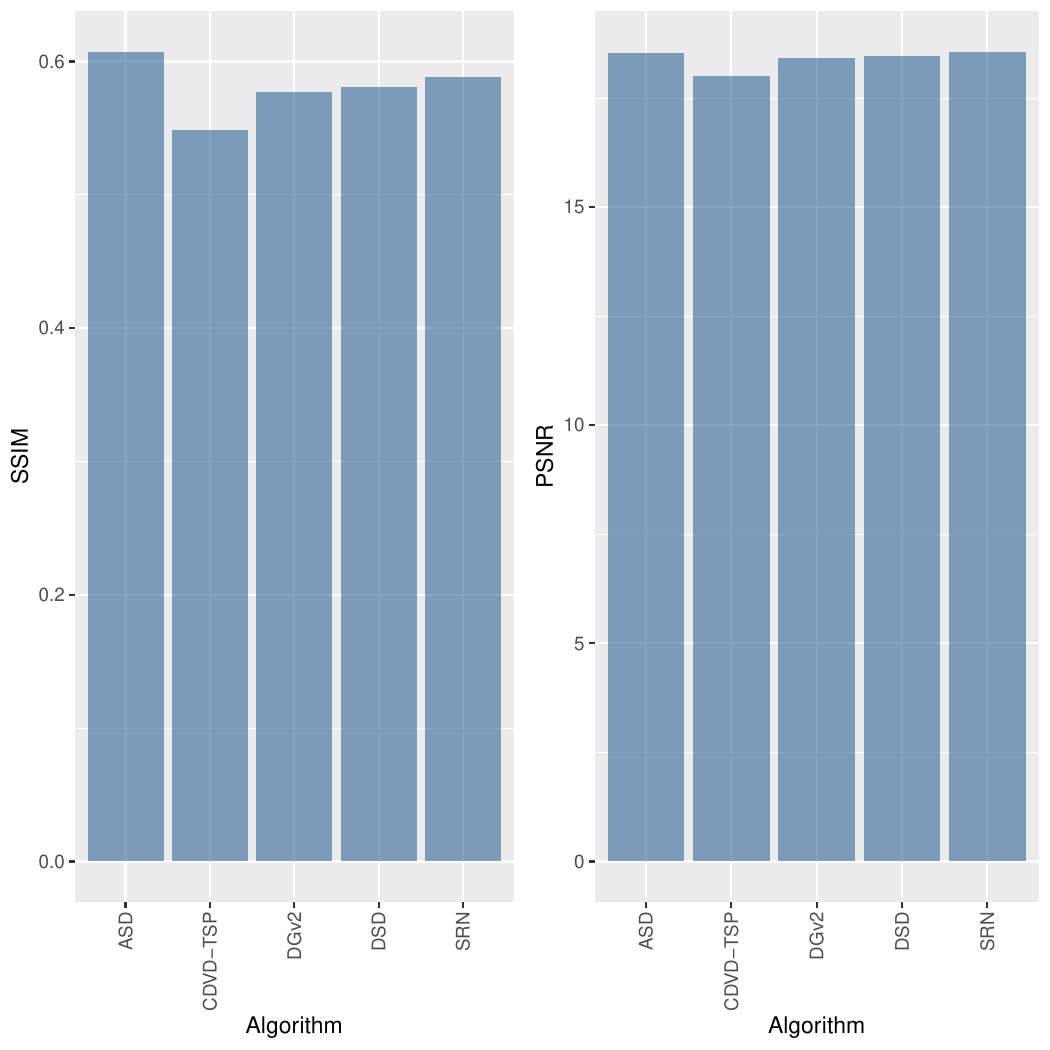}
   \caption{Overall performances (averaged across all stabilization algorithms).}
   \label{fig:pdl1}
  \end{subfigure}
  \hfill
  \begin{subfigure}{0.32\textwidth}
   \centering
   \includegraphics[width=\textwidth]{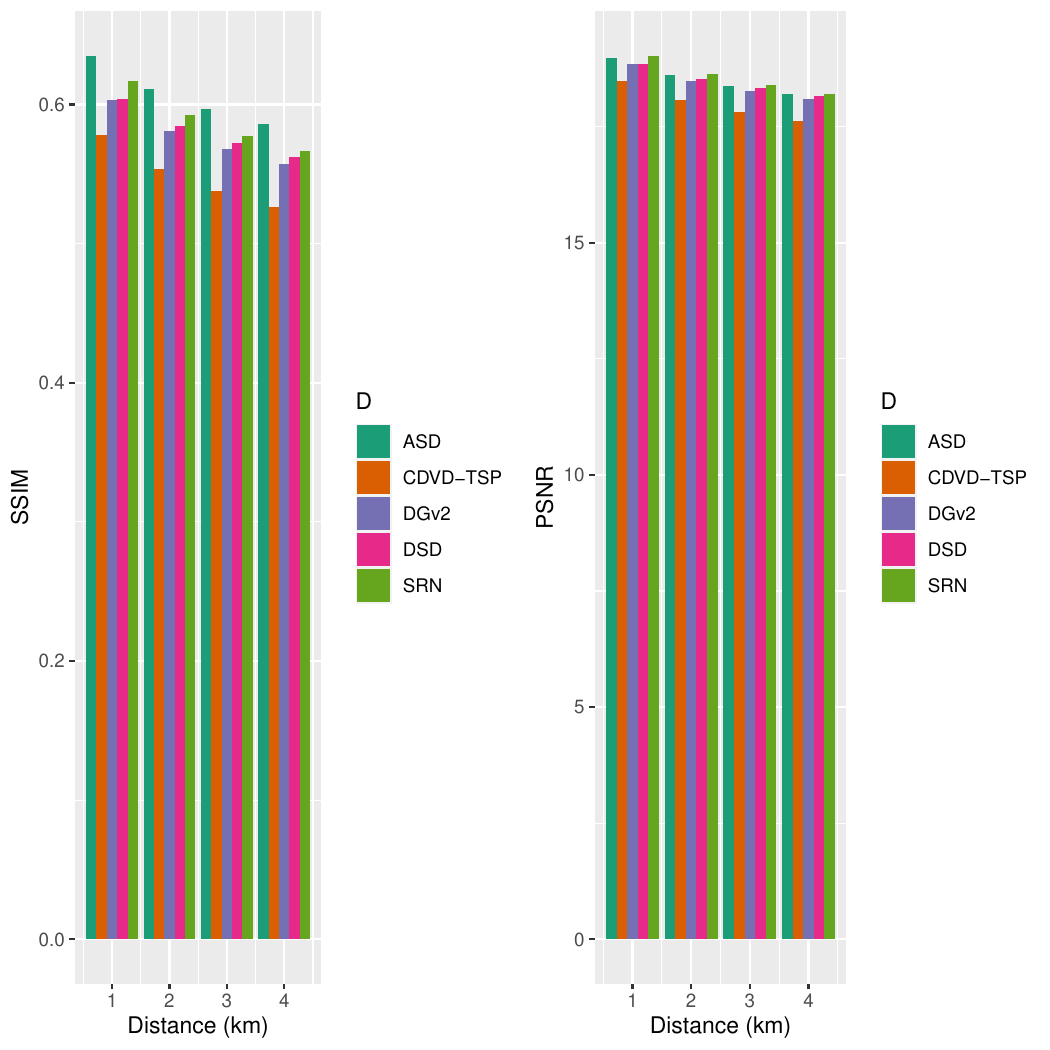}
   \caption{Performance with respect to the sensor-target distance $L$.}
   \label{fig:pdl2}
  \end{subfigure}
   \hfill
   \begin{subfigure}{0.32\textwidth}
    \centering
    \includegraphics[width=\textwidth]{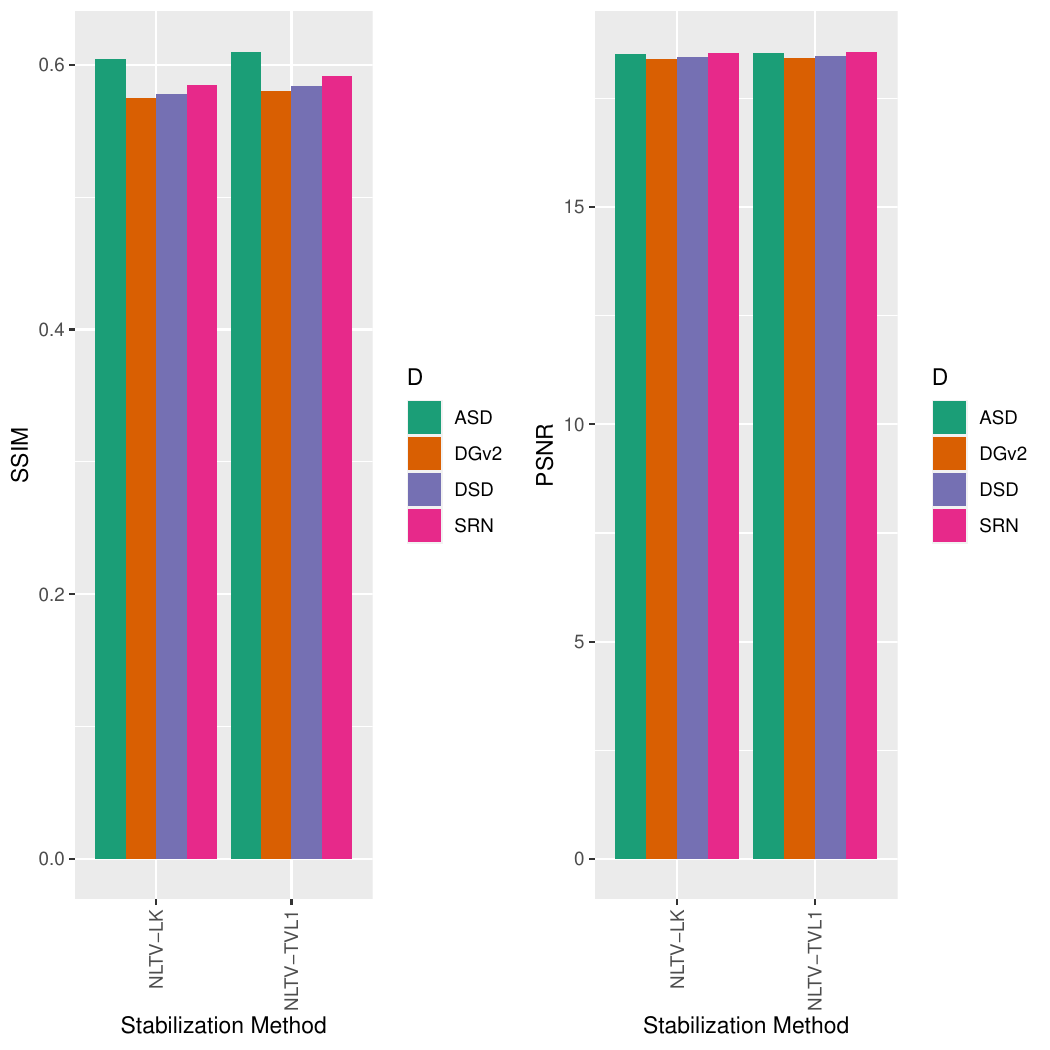}
    \caption{Performance with respect to stabilization.}
    \label{fig:pdl3}
   \end{subfigure}
  \vspace{0.3in}
  
  \begin{subfigure}{0.48\textwidth}
   \centering
   \includegraphics[width=\textwidth]{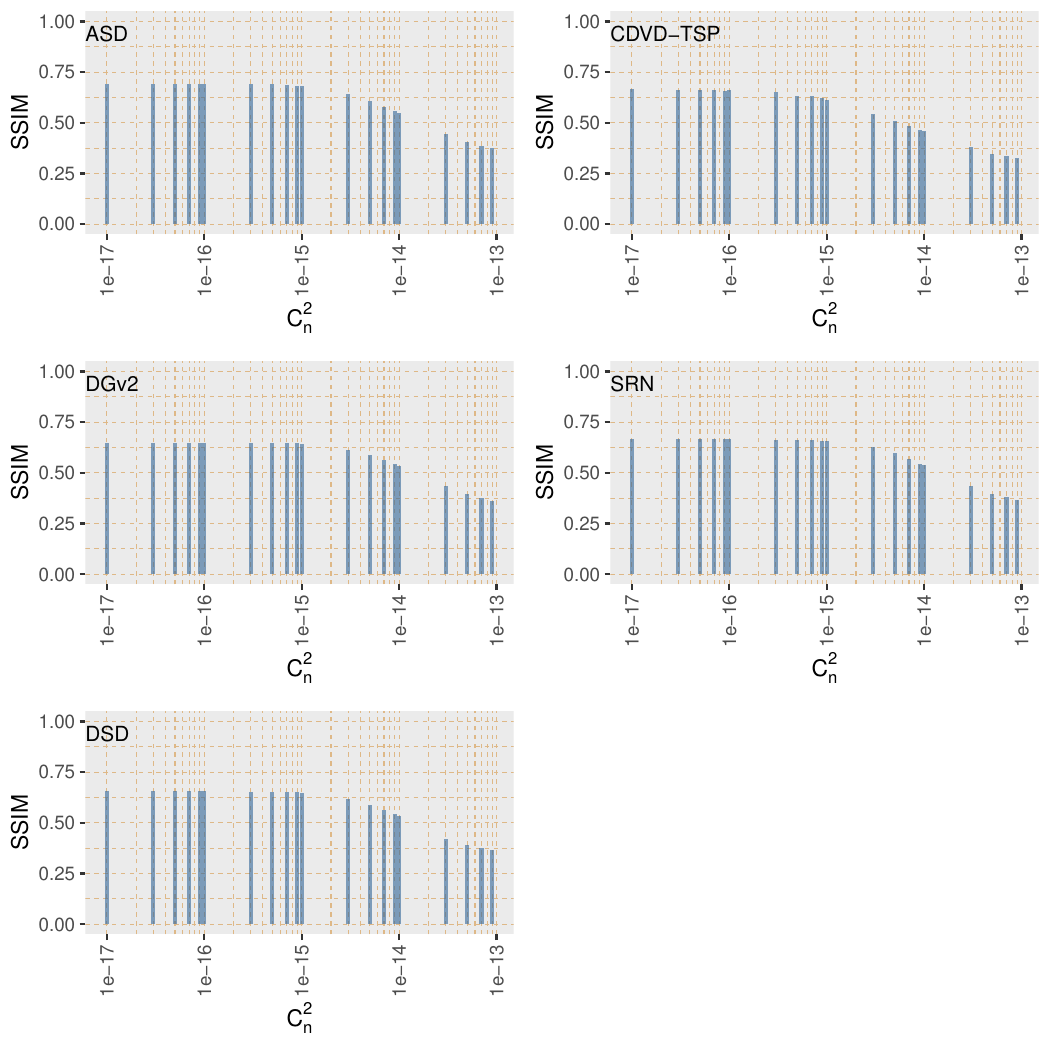}
   \caption{Performance (SSIM) with respect to turbulence strength ($C_n^2$).}
   \label{fig:pdl4}
  \end{subfigure}
  \hfill
  \begin{subfigure}{0.48\textwidth}
   \centering
   \includegraphics[width=\textwidth]{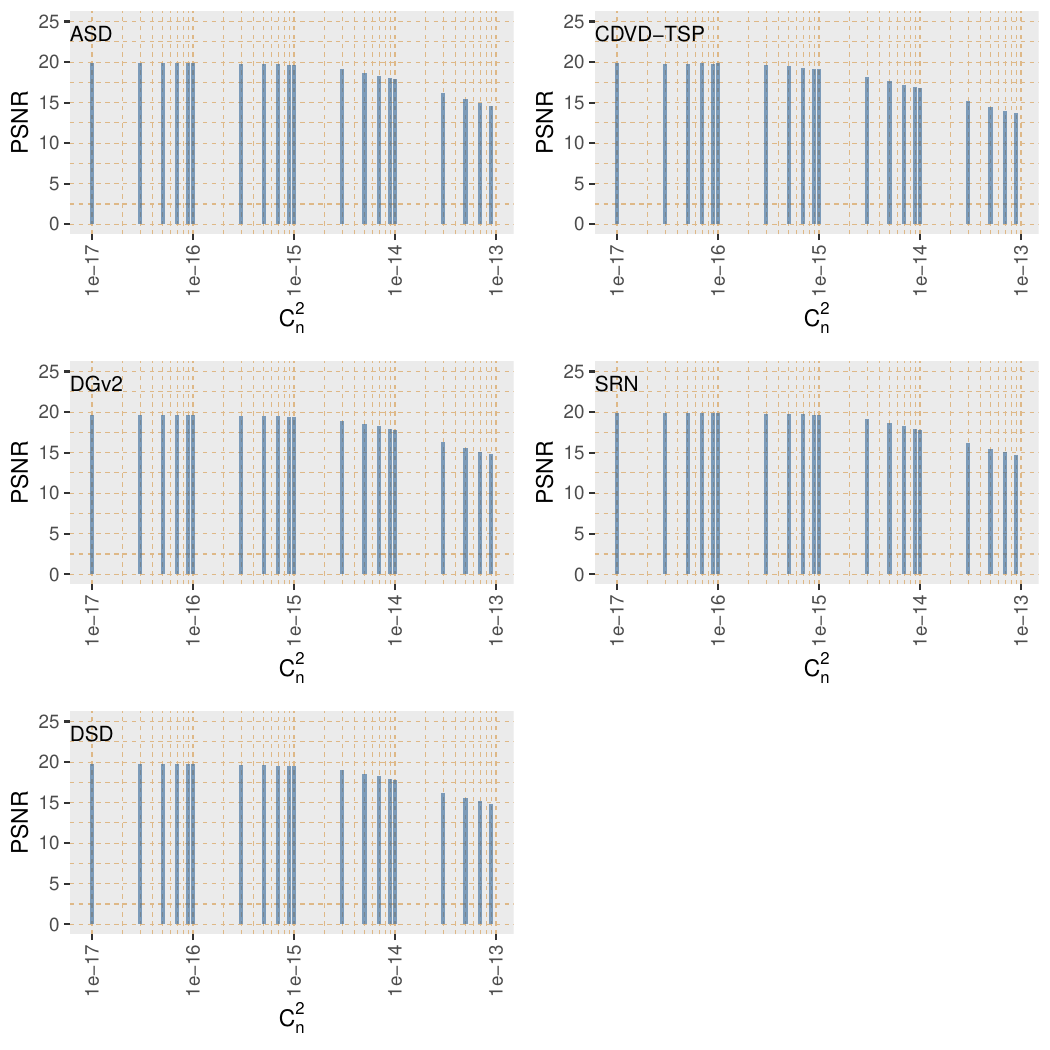}
   \caption{Performance (PSNR) with respect to turbulence strength ($C_n^2$).}
   \label{fig:pdl5}
  \end{subfigure}
  \vspace{0.3in}
  \caption{Performances results for Pre-trained Deep Learning algorithms.}
  \label{fig:pdl}
  \end{figure}

We evaluate the performance of the five selected neural network models mentioned in Section~\ref{sec:deep} on the SOTIS dataset. Unlike the more popular GoPro dataset that was created by averaging a sequence of frames and geared towards simulating motion-blur, the SOTIS dataset contains images that are simulated using real physics and is geared towards providing realistic atmospheric turbulence scenarios. Thus by using the SOTIS dataset, we wish to observe how well the networks adapt to mitigating atmospheric turbulence, instead of removing blur caused by camera shake as originally intended. Because the SOTIS dataset contains a larger number of samples, we perform less epochs for training compared to the original papers. Note that, due to a lack of time, we only ran the evaluations on the $NLTV-LK$ and $NLTV-TVL1$ stabilized sequences. The number of iterations performed by the authors are indicated in Table~\ref{table:parametersPreTraining} and the number of iterations we performed are indicated in Table~\ref{table:parametersReTraining-NLTV-LK} (using $NLTV-LK$), Table~\ref{table:parametersReTraining-NLTV-TVL1} (using $NLTV-TVL1$) and Table~\ref{table:parametersReTraining-CDVD-TSP}. Except for the number of epochs, we used default settings for all models as provided by their respective authors. Another difference compared to the original papers is the number of channels used for the input images to the networks. The networks used RGB images for their training, however since SOTIS only contains grayscale images, we converted the grayscale images to RGB images by copying the same values across all three channels. However, we noticed that such conversion caused convergence problems for ASD using $NLTV-LK$. To circumvent this issue, we stopped the algorithm one step before we detected a sign of divergence.\\
We first evaluate the testing set on the pre-trained networks provided by their respective authors. Because the SOTIS dataset is significantly different from the GoPro dataset, we re-train all networks using SOTIS and present how well re-training compares against the pre-trained network parameters. Using pre-trained parameters would make sense if we expect the probability distribution of the blur kernel for motion deblurring to be comparable to the blur kernel causing atmospheric turbulence. Comparing the results from Figure \ref{fig:pdl} and Figure \ref{fig:dl}, we observe that re-training significantly improves the quality of the estimated image.

Second, we split the SOTIS dataset for training and testing in a 75:25 ratio. We use the sequence of blury frames to train CDVD-TSP, while we feed the stabilized images to all other deblurring networks.

\begin{table}[H]
  \centering
  \caption{Parameters Used for Pre-Training Evaluation}
  \begin{tabular}{|c||c|c|c|c|c|} \hline
    Method & Batch Size  & Steps & Epochs & Training Examples & Total \\ \hline \hline
    DSD & 20 & 200000 & 1887 & 2103 & 4000000 \\ \hline
    DGv2 & 5 & 600000 & 300 & 10000 & 3000000 \\ \hline
    ASD & 4 & 1100000 & 105 & 42000 & 4400000 \\ \hline
    SRN & 16 & 264000 & 2000 & 2103 & 4206000 \\ \hline
    CDVD-TSP & 8 & 381500 & 500 & 6100 & 3050000 \\ \hline
  \end{tabular}
  \label{table:parametersPreTraining}
\end{table}
\begin{table}[H]
  \centering
  \caption{Parameters Used for Re-Training Evaluation NLTV-LK}
  \begin{tabular}{|c||c|c|c|c|c|} \hline
    Method & Batch Size  & Steps & Epochs & Training Examples & Total \\ \hline \hline
    DSD & 14 & 200000 & 217 & 12900 & 2800000 \\ \hline
    DGv2 & 5 & 288960 & 112 & 12900 & 1444800 \\ \hline
    ASD & 6 & 120400 & 56 & 12900 & 722400 \\ \hline
    SRN & 10 & 983000 & 763 & 12900 & 9830000 \\ \hline
  \end{tabular}
  \label{table:parametersReTraining-NLTV-LK}
\end{table}
\begin{table}[H]
  \centering
  \caption{Parameters Used for Re-Training Evaluation NLTV-TVL1}
  \begin{tabular}{|c||c|c|c|c|c|} \hline
    Method & Batch Size  & Steps & Epochs & Training Examples & Total \\ \hline \hline
    DSD & 14 & 187500 & 204 & 12900 & 2625000 \\ \hline
    DGv2 & 5 & 774000 & 300 & 12900 & 3870000 \\ \hline
    ASD & 6 & 290250 & 135 & 12900 & 1741500 \\ \hline
    SRN & 10 & 1022000 & 793 & 12900 & 10220000 \\ \hline
  \end{tabular}
  \label{table:parametersReTraining-NLTV-TVL1}
\end{table}
\begin{table}[H]
  \centering
  \caption{Parameters Used for Re-Training Evaluation CDVD-TSP}
  \begin{tabular}{|c||c|c|c|c|c|} \hline
    Method & Batch Size  & Steps & Epochs & Training Examples & Total \\ \hline \hline
    CDVD-TSP & 1 & 4515000 & 7 & 645000 & 4515000 \\ \hline
  \end{tabular}
  \label{table:parametersReTraining-CDVD-TSP}
\end{table}

The performances of each algorithms are very different from each other for re-training. Looking at the overall performance in Figure~\ref{fig:dl1}, we see that ASD performed much better than the rest of the algorithms. We notice that DGv2 performs better than SRN, whereas the results in ~\citenum{KOH2021103134} suggested that SRN performed better at mitigating motion blur. This seems to indicate that the task of atmospheric turbulence mitigation is indeed different from removing a camera shake induced blur. The overall performance of all re-trained networks shows better results than non-deep learning algorithms. This recommends that investigating the use of neural networks for atmospheric turbulence mitigation could lead to better algorithms than the current state of the art.\\
The difference in the choice of stabilization method for pre-training is less significant as shown in Figure~\ref{fig:pdl3} compared to that for re-training in Figure~\ref{fig:dl3}. However, $NLTV-TVLK$ performs much better for SRN after retraining.\\
The performances of all networks decrease with respect to distance as indicated in Figure~\ref{fig:dl2} (for re-training) and Figure~\ref{fig:pdl2} (for pre-training). A similar decline in performance can also be seen with respect to $C_n^2$ in Figures~\ref{fig:dl4} and \ref{fig:dl5} (for re-training) and in Figures~\ref{fig:pdl4} and \ref{fig:pdl5} (for pre-training).

The observed performances of CDVD-TSP indicate that either a separate stabilization stage is needed, or specific neural networks based stabilization architectures must be developed. For instance, end-to-end training could be improved by potentially incorporating some of the traditional stabilization algorithms into the network. Although our experiments are able to tell how the networks compare against each other, experiments could be performed to study which aspect of the networks contribute to its deblurring power. That could involve removing aspects of the network and seeing how much the performance deteriorates. This could further be used to create networks that are more efficient to tackle atmospheric turbulence in images.

\section{Conclusion}\label{sec:conc}
In this article, we have introduced a new publicly available large dataset, SOTIS, intended to provide simulated sequences impacted by different scenarios of atmospheric turbulence. It also provides the corresponding ground-truth images which can be used for both performing qualitative evaluations of mitigation algorithms, as well as to train neural network architectures for future research. We have presented a first set of evaluation results in both the non-deep learning and deep learning cases. We observe that such systematic evaluation permits to better understand the importance and role of each stage in the restoration process.\\
In future work, we expect to run many more algorithm evaluations to have a clear picture of today's achievements. We also plan to use the learned knowledge to develop more specific deep learning based architectures.

\acknowledgments
The work presented in this paper has been sponsored by the Air Force Office of Scientific Research under the grant number FA9550-21-1-0275.

\appendix    

\bibliography{report} 
\bibliographystyle{spiebib} 
\end{document}